\documentclass{article}

\usepackage{iclr2027_conference,times}

\usepackage{wrapfig}
\usepackage{float}
\usepackage[utf8]{inputenc}
\usepackage[T1]{fontenc}
\usepackage{hyperref}
\hypersetup{hidelinks}
\usepackage{url}
\usepackage{booktabs}
\usepackage{amsmath, amsfonts, amssymb}
\usepackage{graphicx}
\usepackage{natbib}
\usepackage{tabularx}
\usepackage{adjustbox}
\usepackage{placeins}
\usepackage{pifont}
\usepackage{ragged2e}
\usepackage{multirow}
\usepackage[most]{tcolorbox}
\usepackage{xcolor}
\usepackage{tcolorbox}
\usepackage{colortbl}
\usepackage{arydshln}

\definecolor{EditHeatBlue}{HTML}{5C84A6}
\newcommand{\eh}[2]{\cellcolor{EditHeatBlue!#1}#2}

\newcommand{\graydashline}{%
    \arrayrulecolor{black!45}%
    \hdashline[0.4pt/1.4pt]%
    \arrayrulecolor{black}%
}

\iclrfinalcopy

\title{Charts Are Beyond Pixels: Probing for Layer-Wise Chart Understanding and Editing
}

\author{
  Xiaochuan Zhong$^{1}$ \quad
  Yifan Hou$^{2}$ \quad
  Chenxi Pang$^{3}$ \quad
  Shaobo Cui$^{1}$\thanks{Corresponding author.}
  \\
  $^{1}$ DeepDelta Lab, School of Artificial Intelligence, Shanghai Jiao Tong University\\
  $^{2}$ ETH Zurich \quad
  $^{3}$ Google DeepMind
}

\usepackage{xspace}
\newcommand{\benchmarkName}[0]{LayerWiseBench\xspace}

\begin{document}
\maketitle
\raggedbottom

\lhead{Research Manuscript}

\begin{abstract}
Charts are structured visual compositions whose elements have distinct functional roles, semantic correspondences, and visibility relations. This structural view motivates evaluating whether models can understand and manipulate charts at the layer level. Existing chart benchmarks, however, primarily assess the correctness or fidelity of final outputs and do not directly evaluate these layer-wise behaviors. We present LayerWiseBench, a benchmark organized around three core concepts, layer attribution, layer binding, and visibility ordering, that structure its chart-understanding and chart-editing evaluations. Generated from executable chart programs, LayerWiseBench pairs each rendered chart with spatially aligned per-layer RGBA assets and construction-derived labels for functional roles, semantic bindings, and visibility relations. From this layer-wise representation, we derive controlled understanding questions, editing targets, reference images, and evaluation regions. It contains 2,800 source charts across 14 chart paradigms, from which we derive 7,329 layer-wise understanding questions and 53,791 instruction-guided editing variants. Among the evaluated VLMs, Qwen3.5-27B, which achieves the highest QA macro-average, obtains 93.04\% accuracy on layer attribution and 97.46\% on layer binding, but only 61.46\% on visibility ordering. Across the four evaluated image editors, overall mIoU ranges from 1.49\% to 4.93\%, and visibility-constrained edits have the lowest mIoU for every editor, ranging from 0.37\% to 2.00\%. Taken together, these results identify tasks involving front-to-back relations between overlapping components as a recurring challenge across understanding and editing, motivating more explicit modeling of component identity and visibility relations.
\end{abstract}

\section{Introduction}

Charts are structured visual compositions rather than merely rendered pixel arrays. \citet{wilkinson2005grammar} describes a statistical graphic through components such as data, transformations, scales, graphical elements, coordinate systems, and guides. Wickham further develops this view into a layered grammar, in which a plot is constructed from one or more layers together with scales, coordinates, and facets \citep{wickham2010layered}. Similarly, \citet{satyanarayan2017vegalite} operationalized a related compositional view through declarative specifications of data transformations, marks, encodings, and layered or multi-view compositions. Taken together, these frameworks provide a compositional account of charts in terms of graphical components, mappings, guides, and view structures. This view motivates our study of how models for chart understanding and editing can be evaluated at the layer level.

Established chart benchmarks cover tasks such as question answering, code generation, and editing, often assessing task-level outputs through answer correctness, code execution, or fidelity of rendered charts \citep{masry2022chartqa,wu2025plot2code,zhao2025chartedit}. More recent work incorporates structured signals through visual grounding and scene-graph comparison \citep{vogel2025refchartqa,goswami2025charteval}. We build on these structured signals by organizing evaluation around functional layers and separately assessing attribution, binding, and visibility.

\begin{figure*}[t]
    \centering
    \includegraphics[width=0.94\textwidth]
    {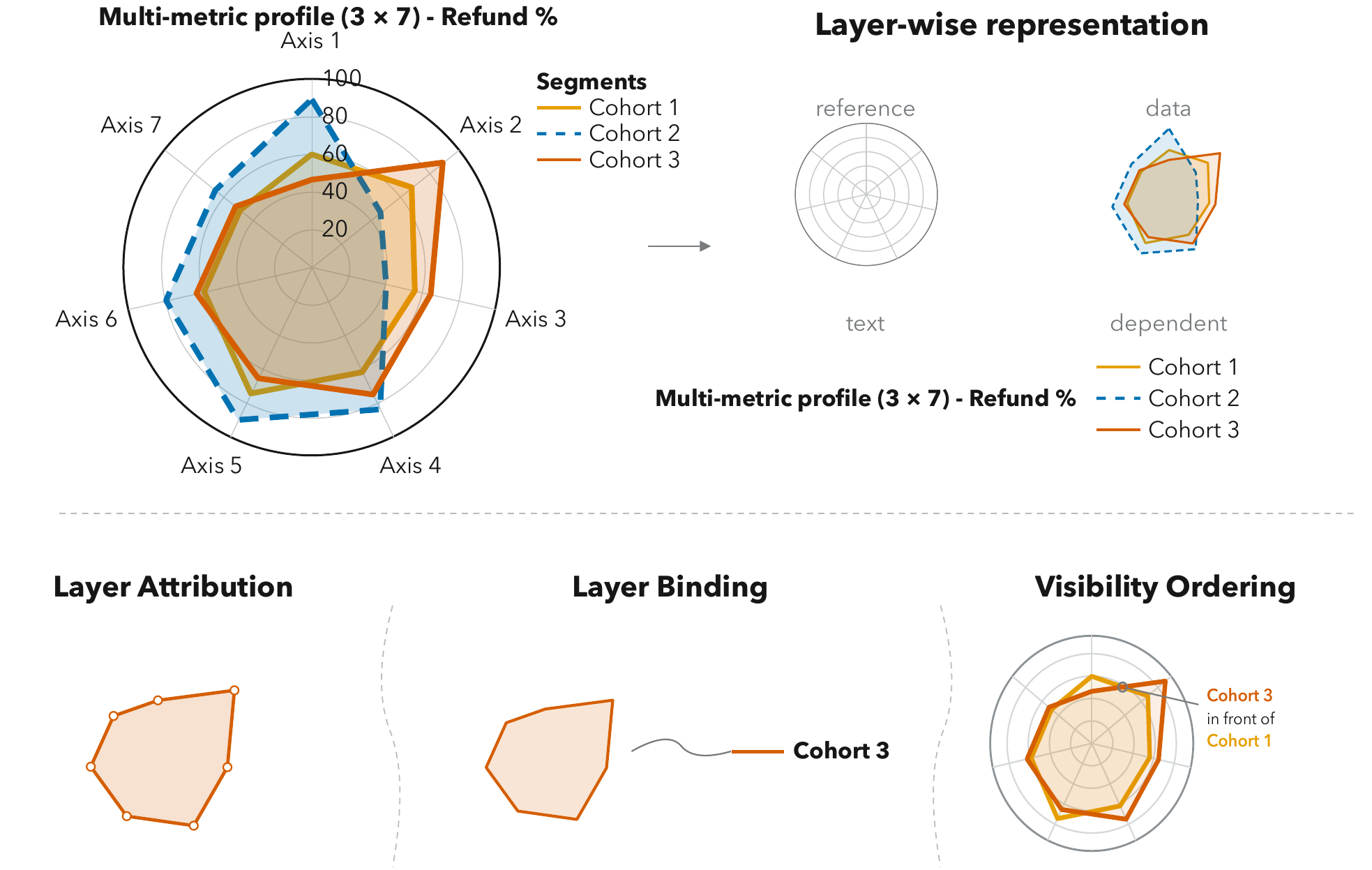}
    \caption{
    Overview of the layer-wise representation and the three dimensions evaluated in \benchmarkName.
    \textbf{Left:} A rendered radar chart.
    \textbf{Upper right:} The corresponding representation separates its reference, data, text, and dependent components.
    \textbf{Bottom:} Layer attribution captures the elements and properties within a layer; layer binding captures semantic correspondences across layers; and visibility ordering records the front-to-back relation between overlapping components.
    }
    \label{fig:intro_teaser}
\end{figure*}

To make this layered grammar explicit for evaluation, we formulate a layer-wise representation for evaluation informed by visualization grammars
\citep{wilkinson2005grammar,wickham2010layered,mackinlay1986automating,satyanarayan2017vegalite}. Figure~\ref{fig:intro_teaser} illustrates this representation by separating a rendered chart into reference, data, text, and dependent components. The representation captures the elements and semantic and visual properties within a layer, together with semantic correspondences across layers, following established attribute--relation distinctions in structured visual evaluation and chart analysis \citep{johnson2015scenegraphs,krishna2017visualgenome,zhao2022vlchecklist,siegel2016figureseer,goswami2025charteval}. Because these properties and correspondences do not fully determine the appearance of overlapping components, the representation additionally records their front-to-back order \citep{porter1984compositing,snyder1998visibility,lee2022instancewise}. These three aspects give rise to our core concepts corresponding respectively to the elements and properties within each layer, semantic correspondences across layers, and the front-to-back relation among overlapping components.

Building on these concepts, we introduce \emph{\benchmarkName}, a benchmark that evaluates model behavior with respect to this layered grammar through two tracks: \emph{layer-wise understanding}, formulated as chart questions, and \emph{layer-wise editing}, formulated as targeted chart editing tasks. Built from executable chart programs, \benchmarkName uses construction records and spatially aligned layer renderings to derive annotations of functional roles, semantic correspondences, and visibility relations. Overall, \benchmarkName contains \emph{2,800} source charts across \emph{14} chart paradigms, from which we derive \emph{7,329} layer-wise understanding questions and \emph{53,791} layer-wise editing variants.

On layer-wise understanding, Qwen3.5-27B, which achieves the highest macro-average accuracy, reaches 93.04\% on layer attribution and 97.46\% on layer binding, but drops to 61.46\% on visibility ordering. Visibility ordering is the lowest-scoring dimension for all seven VLMs in the main comparison. For image editing, overall mIoU ranges from 1.49\% to 4.93\% across the four evaluated editors, and visibility-constrained edits have the lowest mIoU for every editor, ranging from 0.37\% to 2.00\%. Across both tracks, these results identify tasks involving front-to-back relations between overlapping components as a recurring challenge.

Our contributions are:
\begin{itemize}

    \item We introduce \benchmarkName, a benchmark with layer-wise chart understanding and editing tracks, built from executable templates that register task-relevant graphical elements into functional layer groups. These registrations link each template's construction code to its functional layer structure, yielding construction-grounded annotations and spatially aligned layer renders for both tracks.

    \item We evaluate nine vision-language models and four image editing models on their respective tracks. Visibility ordering is the lowest-scoring understanding dimension for all seven VLMs in the main comparison, and visibility-constrained edits have the lowest mIoU for all four image editing models.
    
\end{itemize}

\section{Related Work}

\subsection{Chart Understanding Patterns and Benchmarks}

Chart understanding methods broadly follow two patterns. End-to-end approaches directly map a rendered chart to an answer or textual output, whereas other approaches first recover an intermediate representation, such as detected chart components, a data table, rendering code, or structured triples, before downstream reasoning. Representative systems integrate chart derendering with comprehension, translate plots into tables for language-model reasoning, or explicitly separate chart perception from structured reasoning \citep{liu2023deplot,cheng2023chartreader,liu2023matcha,xia2023structchart}. These works show that intermediate chart structure is a recurring and practically useful object of modeling.

Chart understanding benchmarks have progressively broadened their coverage of questions, chart types, and reasoning skills. ChartQA, MMC-Benchmark, ChartX, ChartBench, and ChartQAPro extend evaluation toward human-written questions, multi-task reasoning, diverse chart forms, and real-world settings \citep{masry2022chartqa,liu2024mmc,xia2024chartx,xu2024chartbench,
masry2025chartqapro}. RefChartQA further connects answers to supporting chart elements through visual grounding \citep{vogel2025refchartqa}. More recent benchmarks extend this trajectory toward spatial chart-element localization, fine-grained multi-target grounding, and infographic charts \citep{liu2026start,niu2026chartreg,li2026chartgalaxy}. These advances substantially broaden the scope of chart understanding evaluation. Nevertheless, their benchmark formulations generally do not make the functional organization of chart components into layers an explicit organizing principle. \benchmarkName complements these efforts by centering evaluation of chart understanding on this layered organization.

\subsection{Chart Generation Patterns and Editing Benchmarks}

Chart generation and editing follow two broad paradigms. In one, a model first recovers or modifies a structured chart specification and then renders the result, following earlier visualization reverse-engineering work that reconstructs visual encodings from chart images \citep{poco2017reverse}. In the other, a model directly transforms a rendered image according to a natural-language instruction, following the paradigm of instruction-guided image editing \citep{brooks2023instructpix2pix}. This distinction shapes the signals available for evaluation: structured specifications expose executability and symbolic structure, whereas outputs from direct image editing are typically assessed through the resulting visual transformation.

Existing benchmarks instantiate these paradigms with different evaluation targets. Plot2Code, ChartMimic, and ChartAnchor evaluate chart reconstruction through executable code, rendered similarity, or recovered data \citep{wu2025plot2code,yang2025chartmimic,li2026chartanchor}. ChartEdit, ChartM3, ChartEditVista, and ChartEditBench evaluate code-oriented editing under natural-language, multimodal, image-conditioned, or multi-turn instructions \citep{zhao2025chartedit,yang2025chartm3,chen2026charteditor,kapadnis2026charteditbench}. FigEdit and ChartE$^{3}$ evaluate end-to-end image editing through edit fidelity and preservation of the remaining chart \citep{li2026charts,li2026charte3}, while \citet{yu2026chartsync} specifically tested whether textual and geometric changes remain synchronized under cascading edits. \citet{goswami2025charteval} instead assess final chart quality through hierarchical scene-graph similarity. Together, these benchmarks cover reconstruction quality, instruction following, multi-turn interaction, and dependency propagation. Within this line of work, \citet{yu2026chartsync} studied coordinated text-to-geometry changes, which is closely related to the coordinated updates represented in our Binding-consistent family. The editing track of \benchmarkName also adopts an end-to-end interface, asking models to transform a chart image according to an instruction. Its layer-wise organization comes from construction-derived layer annotations, which group instances into Local target, Binding-consistent, and Visibility-constrained edit families and enable performance to be reported separately across these conditions.

\section{Layer-wise Chart Representation and Formulation}
\label{sec:formulation}

Visualization grammars and component-based scene representations describe charts through components, their properties, and their organization \citep{wilkinson2005grammar,wickham2010layered,satyanarayan2017vegalite,liu2025manipulable}. We therefore represent each chart in our controlled construction domain as

\begin{equation}
\label{eq:layerwise_representation}
S=(C,\Lambda),
\end{equation}

where $C$ is the collection of components instantiated in the chart, and $\Lambda$ records how those components are organized and related. We use component broadly to include visible chart elements and chart-level layout structures together with their resolved data, content, appearance, and spatial properties. The executable source program produces the rendered image and the construction records from which $S$ is instantiated. Within this controlled construction domain, $S$ provides a component-level description of the chart for the evaluated tasks.

Within $\Lambda$, $\lambda$ denotes the functional-layer assignment and $\mathcal{B}$ records cross-layer semantic bindings. For overlapping component pairs, $\mathcal{V}$ records pairwise visibility ordering as a resolved front-to-back relation, since compositing order can change the visible result \citep{porter1984compositing}. The three evaluation dimensions introduced in the Introduction are defined over these structures.

Appendix~\ref{app:terminology} summarizes the terminology used throughout the paper, and Appendix~\ref{app:layer_ontology} gives the complete operational ontology.

\section{\benchmarkName Construction}

\begin{figure*}[t]
    \centering
    \includegraphics[
        width=\linewidth,
    ]{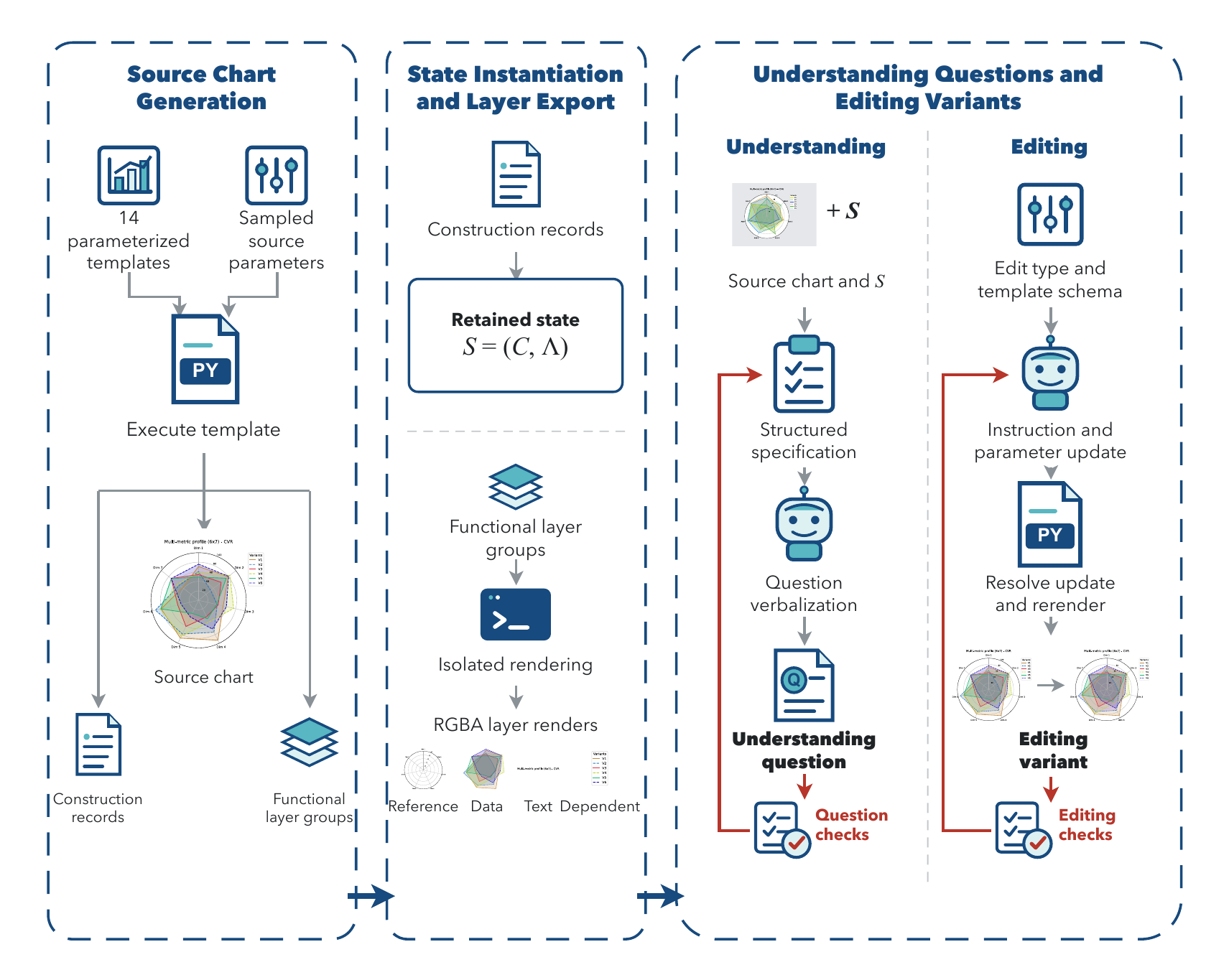}
    \caption{
    Construction pipeline of \benchmarkName. Parameterized templates produce source charts, construction records, and functional layer groups, from which we instantiate \(S=(C,\Lambda)\) and export spatially aligned RGBA layer renders. The resulting representation and renderings are used to generate and verify layer-wise understanding questions and editing variants.
    }
    \label{fig:construction_pipeline}
\end{figure*}

The construction of \benchmarkName follows two principles. \emph{Layer-aware organization} makes component roles and relations explicit, allowing the same component representation to support understanding questions and editing conditions. \emph{Source-based construction} grounds both tracks in executable chart programs: construction records determine the answers to understanding questions, while parameter interventions and rerendering produce reference edits. Figure~\ref{fig:construction_pipeline} summarizes the complete construction process.

\subsection{Source Chart Generation}

We construct a library of source charts designed to expose varied component organizations and relations. A preliminary qualitative analysis of chart-editing failures informed its scope, and the final library comprises 14 paradigms spanning diverse mark, layout, and overlap structures. We fixed its composition before the reported model evaluation.

Each paradigm is implemented as a parameterized template. The same template produces a source chart and its reference edits by updating specified construction parameters while inheriting the remaining source configuration. During rendering, task-relevant graphical elements are registered into functional layer groups, which are rendered on the same canvas to produce spatially aligned transparent RGBA layer renders alongside the composite image. The instantiated graphical elements and construction records provide $C$ and the structures $\lambda$, $\mathcal{B}$, and $\mathcal{V}$ retained in $\Lambda$. Aligned layer renders provide their spatial grounding in the composite image.

We generate 200 source charts per paradigm, yielding 2,800 charts in total. Splits are assigned by source chart: each paradigm contributes 160/20/20 charts to the training, validation, and test sets, respectively, for totals of 2,240/280/280. Every derived question or editing variant inherits the split of its source chart.

\subsection{Layer-wise Understanding Instance Generation}

For each source chart, we derive candidate understanding queries from component properties in $C$ and the structures $\lambda$, $\mathcal{B}$, and $\mathcal{V}$ in $\Lambda$, grounding the selected component or pair in the corresponding composite image. We instantiate a query only when its referent is visually identifiable and its answer is uniquely determined by the retained construction. Visibility ordering queries additionally require verified local overlap between the aligned rendered layers.

Each eligible query is encoded as a structured specification that fixes its family, target, answer choices, and gold answer before wording. An LLM then verbalizes only the chart-specific question stem without altering these fields. Generated questions are checked for consistency with the specification, visual grounding, and answer leakage; failed checks trigger revision or review before finalization.

The understanding branch covers 2,600 source charts across 13 paradigms. Across all splits, it contains 7,329 questions: 3,172 on layer attribution, 3,176 on layer binding, and 981 on visibility ordering. The test split contains 727 questions derived from 260 source charts, with 316, 315, and 96 instances in the three families, respectively.

\subsection{Layer-wise Editing Variant Generation}

Whereas understanding instances query component properties and relations in $S$, editing variants are generated through controlled parameter interventions on the source program and subsequently organized by structural conditions derived from $S$. Each intervention updates a supported target or construction property, and re-executing the same paradigm template yields the corresponding reference edit.

For each supported edit type, an LLM proposes a candidate natural-language instruction together with a structured parameter update constrained by the template schema. Deterministic construction code resolves the update against the source configuration and restricts it to the template's admissible parameter domain. We check the instruction against the resolved update and verify that the resulting parameter changes are reflected in rebuilt renderer profiles. Validated edit specifications are then instantiated over additional admissible parameter values while keeping the target, edit type, and edited parameter set fixed; the same checks are applied to every resulting candidate.

For family-level analysis, we annotate eligible test variants according to their construction-derived structural conditions. A \emph{Local target edit} modifies a specified target or layout property without requiring a relation-specific condition. A \emph{Binding-consistent edit} additionally produces a coordinated visible change in a construction-linked dependent component, whereas a \emph{Visibility-constrained edit} modifies a target within a verified overlap while retaining the resolved front-to-back relation. These conditions may co-occur, but each variant is assigned to one family for reporting.

Across all 14 paradigms, this process generates 53,791 candidate variants. By inheriting the split of their source charts, 43,015, 5,384, and 5,392 candidates belong to the training, validation, and test sets, respectively. The evaluation setting specifies the validated test subset used for scoring.

Further construction and validation details are provided in Appendix~\ref{app:construction_details}.

\section{Evaluation Setting}

\begin{wrapfigure}{r}{0.48\textwidth}
    \centering
    \includegraphics[
        width=\linewidth
    ]{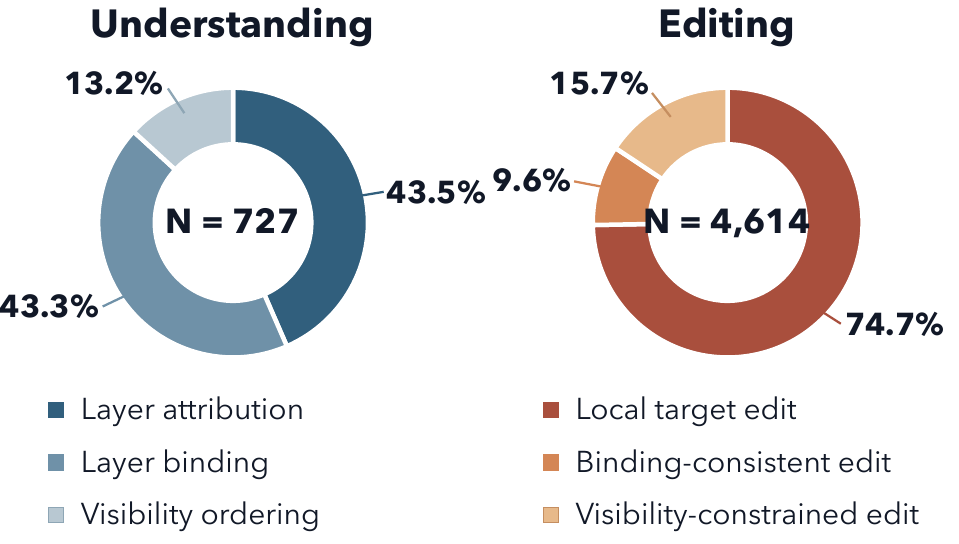}
    \caption{Family composition of the understanding and scored editing test sets.}
    \label{fig:task_family_distribution}
\end{wrapfigure}

\subsection{Layer-wise Understanding Setting}

We evaluate Qwen3.5-0.8B/2B/4B/9B/27B, Qwen3-VL-8B, InternVL3-8B/14B, and MiniCPM-V-4.5~\citep{qwen2026qwen35,bai2025qwen3vl,zhu2025internvl3,yu2025minicpmv45}. Each model receives a rendered chart, a multiple-choice question, its answer choices, and the four functional layer definitions, and returns one textual response. The 727-item test set contains 316 questions on layer attribution, 315 on layer binding, and 96 on visibility ordering.

Responses are mapped to the listed choices, and an output that cannot be mapped is counted as incorrect. All models are evaluated on the full test set. We report accuracy for each task family, their unweighted macro-average, and the rate of invalid responses over all 727 items.

\subsection{Layer-wise Editing Setting}

We compare FLUX Kontext Dev, InstructPix2Pix, OmniGen2, and Qwen Image Edit 2509~\citep{bfl2025fluxkontext,brooks2023instructpix2pix,wu2025omnigen2,qwen2025imageedit2509}. Each system receives only a source chart and an editing instruction. The reference edit and construction annotations are reserved for evaluation. Editing is assessed as an end-to-end image transformation under construction-derived conditions rather than as direct recovery of the formal representation.

The common test set contains 4,614 items with a visible reference change and a construction-derived family assignment. The set comprises 3,447 Local target, 444 Binding-consistent, and 723 Visibility-constrained items (Figure~\ref{fig:task_family_distribution}). All systems are scored on this same set after their outputs are normalized to the canvas of the source chart.

We use mIoU from PaintBench~\citep{xu2026paintbench} as our primary editing metric and compute Edit Accuracy and Preservation Accuracy as complementary diagnostics. The edited region is defined by pixels that differ between the source and reference images. At each color-distance tolerance, mIoU rewards pixels in this region that match the reference and penalizes both incorrect target pixels and deviations elsewhere. The reported mIoU averages over tolerances and items. Edit Accuracy isolates correctness within the edited region, whereas Preservation Accuracy measures how well the remaining image is preserved. All three are percentages, with higher values indicating better performance. Overall and family-level results in the main text are reported with mIoU.

We additionally compute mean SSIM between the output and reference edit over all 4,614 items, scaled by 100, together with mean SSIM between the source chart and reference edit as a no-edit baseline~\citep{wang2004ssim}. Higher SSIM indicates greater image similarity. On a fixed, model-independent subset of 568 content-preserving font and style edits, we report mean OCR Normalized Edit Similarity (OCR-NES) using PP-OCRv6~\citep{zhang2026ppocrv6}. OCR-NES is reported on a 0--100 scale and measures preservation of unchanged text.
Complete scoring definitions and implementation details are provided in Appendix~\ref{app:evaluation_metrics}.

Inference settings for both tracks are provided in Appendix~\ref{app:inference_settings}.

\section{Experiments and Analysis}

\begin{wrapfigure}[15]{r}{0.44\columnwidth}
    \vspace{-8pt}
    \centering
    \includegraphics[
        width=\linewidth,
    ]{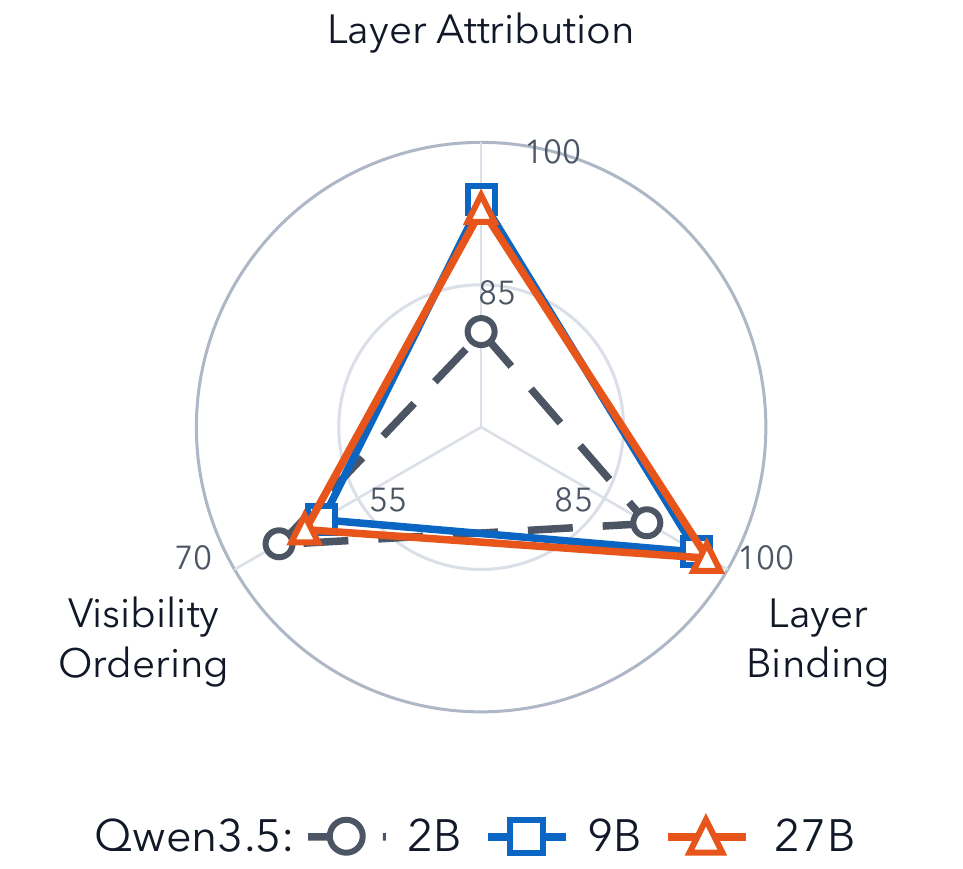}
    \caption{
    Qwen3.5 understanding accuracy. Axes use different ranges.
    }
    \label{fig:understanding_radar}
    \vspace{-8pt}
\end{wrapfigure}

We organize our experiments around two questions:

\begin{itemize}
    \item \textbf{Q1:} Across the three understanding dimensions, which is most often the lowest-scoring for the evaluated VLMs, and what does this pattern reveal about their layer-wise understanding?

    \item \textbf{Q2:} How closely do the evaluated image editors match construction-grounded reference edits as measured by mIoU, and what does the comparison across the three edit families reveal about editing under layer-derived structural conditions?
\end{itemize}

\begin{table*}[!t] 
\caption{ Layer-wise understanding results on the test split. All values are percentages. Macro Avg. is the unweighted average over the three accuracies for the three task families. Invalid denotes the percentage of outputs that cannot be parsed into a valid answer. Invalid outputs are counted as incorrect in the family accuracies and are reported separately. The main comparison uses a 5\% ceiling on image+text invalid choice rates, with results for Qwen3.5-0.8B and 4B reported in Appendix~\ref{app:additional_understanding_results}.} 
\label{tab:qa_main} 
\centering 
\small 
\setlength{\tabcolsep}{4pt} 
\renewcommand{\arraystretch}{1.05} 
\resizebox{\linewidth}{!}{ 
\begin{tabular}{lrrrrr} 
\toprule 
\textbf{Model} & 
\textbf{Macro Avg.} $\uparrow$ & 
\textbf{Layer Attr.} $\uparrow$ & 
\textbf{Layer Bind.} $\uparrow$ & 
\textbf{Visibility Order.} $\uparrow$ & 
\textbf{Invalid} $\downarrow$ \\ 
\midrule 
Qwen3.5-27B & \textbf{83.99} & 93.04 & \textbf{97.46} & 61.46 & \textbf{0.00} \\
Qwen3.5-9B & 83.18 & 93.99 & 96.19 & 59.38 & \textbf{0.00} \\
Qwen3.5-2B & 78.27 & 80.06 & 90.16 & \textbf{64.58} & 1.65 \\
InternVL3-14B & 80.51 & \textbf{94.30} & 88.89 & 58.33 & 0.55 \\
InternVL3-8B & 70.52 & 79.75 & 75.56 & 56.25 & 3.30 \\
Qwen3-VL-8B & 78.37 & 90.51 & 94.60 & 50.00 & \textbf{0.00} \\
MiniCPM-V-4.5 & 80.39 & 93.35 & 86.35 & 61.46 & \textbf{0.00} \\
\bottomrule 
\end{tabular} 
} 

\end{table*}

\FloatBarrier

\subsection{Layer-wise Understanding Experiments}

\paragraph{Visibility ordering is the lowest-performing task family.}

Visibility ordering has the lowest accuracy for all seven VLMs in the main comparison. Even the best result reaches only 64.58\% on 96 binary questions, compared with 50\% accuracy under uniform random guessing. These results indicate that the evaluated VLMs still struggle to resolve front-to-back relations between overlapping chart components.

Text-only accuracy on layer attribution remains 78.48--92.41\%, indicating that many questions in this family can be answered without the chart image (Appendix~\ref{app:additional_understanding_results}).

\paragraph{Higher average accuracy does not imply better visibility ordering.}

Macro-average accuracy increases with model size among the Qwen3.5 and InternVL3 checkpoints in the main comparison (Table~\ref{tab:qa_main}). Qwen3.5 improves from 78.27\% at 2B to 83.18\% at 9B and 83.99\% at 27B, while InternVL3 improves from 70.52\% at 8B to 80.51\% at 14B. The gain from Qwen3.5-9B to 27B is smaller than that from 2B to 9B.

Visibility ordering follows a different pattern in Qwen3.5. The 2B model scores 64.58\%, compared with 59.38\% for 9B and 61.46\% for 27B (Figure~\ref{fig:understanding_radar}). Higher aggregate accuracy can therefore coexist with a persistent weakness in resolving overlaps, motivating explicit supervision of visibility ordering as a complement to increasing model size.

\begin{table*}[t]
\centering
\small
\caption{
Editing performance measured by mIoU.
All values are percentages; higher is better.
Overall is the item-weighted mean over all 4,614 test items.
Darker shading indicates higher mIoU on a common scale; row maxima are bold.
}
\label{tab:editing_results}
\setlength{\tabcolsep}{4pt}
\renewcommand{\arraystretch}{1.05}

\begin{tabular}{@{}l r c c c c@{}}
\toprule
\textbf{Editing family}
& \textbf{$N$}
& \shortstack{\textbf{FLUX}\\\textbf{Kontext Dev}}
& \shortstack{\textbf{Instruct}\\\textbf{Pix2Pix}}
& \textbf{OmniGen2}
& \shortstack{\textbf{Qwen Image}\\\textbf{Edit 2509}} \\
\midrule

Local target edit
& 3,447
& \eh{40}{\textbf{6.22}}
& \eh{16}{1.77}
& \eh{37}{5.60}
& \eh{33}{4.86} \\

Binding-consistent edit
& 444
& \eh{14}{1.52}
& \eh{12}{1.14}
& \eh{20}{2.55}
& \eh{24}{\textbf{3.22}} \\

Visibility-constrained edit
& 723
& \eh{11}{0.90}
& \eh{8}{0.37}
& \eh{14}{1.50}
& \eh{17}{\textbf{2.00}} \\
\midrule

Overall
& 4,614
& \eh{33}{\textbf{4.93}}
& \eh{14}{1.49}
& \eh{32}{4.67}
& \eh{29}{4.26} \\
\bottomrule
\end{tabular}
\end{table*}
\subsection{Layer-wise Editing Results}

\paragraph{Pixel-level agreement with the reference remains low.}

Table~\ref{tab:editing_results} reports the overall and family-level mIoU results. Across the 4,614 evaluation items, overall mIoU ranges from 1.49\% to 4.93\%. Because mIoU evaluates the target and preservation regions jointly, these scores indicate that the evaluated editors do not yet combine pixel-accurate target modification with preservation of the surrounding chart.

\paragraph{Visibility-constrained edits receive the lowest mIoU.}

mIoU ranges from 1.77\% to 6.22\% for local target edits, from 1.14\% to 3.22\% for binding-consistent edits, and from 0.37\% to 2.00\% for visibility-constrained edits. The visibility-constrained subset is the lowest-scoring editing family for all four evaluated editors. Current editors therefore reproduce reference edits less accurately when the target participates in a visible overlap and the original front-to-back relation must be preserved.

\begin{figure*}[t]
    \centering
    \includegraphics[width=\linewidth]{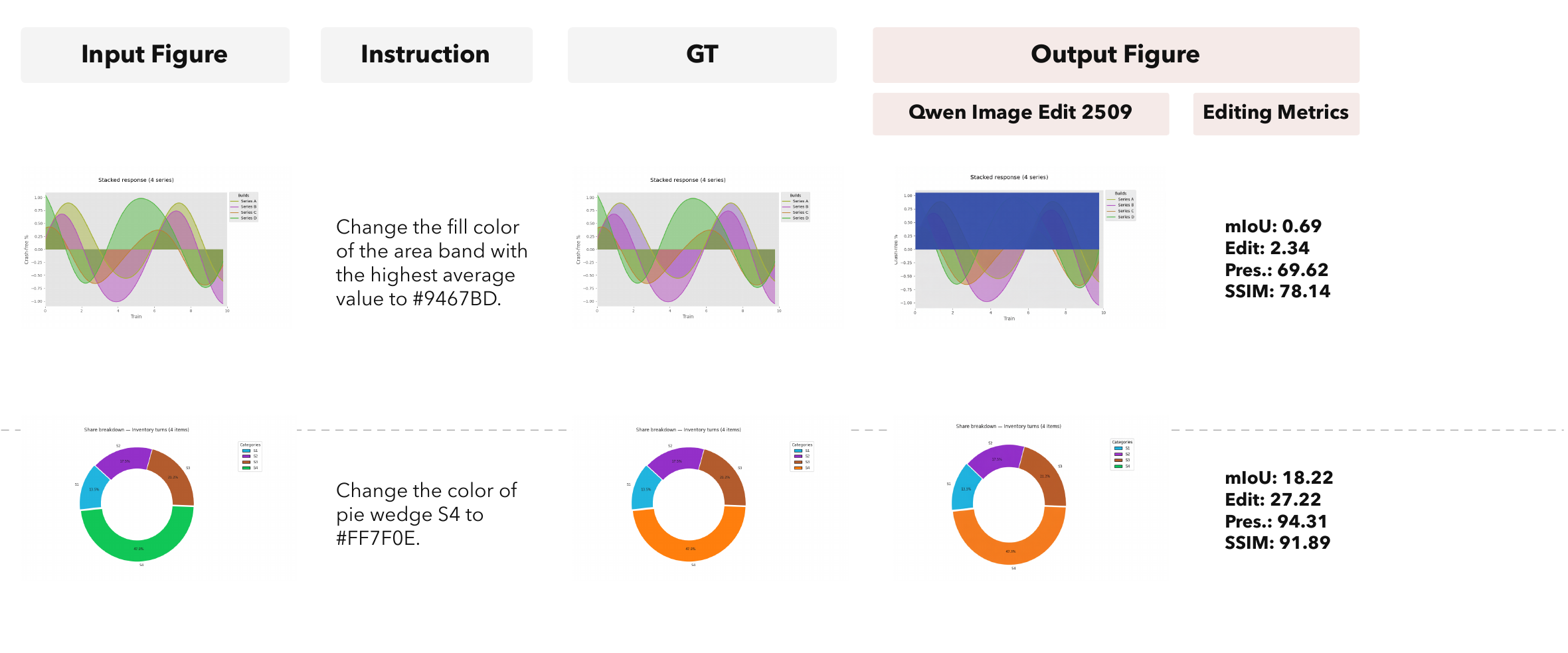}
    \caption{
    Qualitative editing examples from Qwen Image Edit 2509. The area output changes a broad non-target region, whereas the pie output more closely matches the reference edit.
    }
    \label{fig:editing_metric_examples}
\end{figure*}

\paragraph{Qualitative examples.}
Figure~\ref{fig:editing_metric_examples} compares two outputs from Qwen Image Edit 2509 with their reference edits. Their metric values reflect the contrast between broad unintended changes and closer agreement with the reference.

Taken together, the results identify visibility-related cases as the clearest shared weakness across the two tracks. Appendix~\ref{app:additional_results} reports complementary understanding controls and editing metrics. Appendix~\ref{app:visibility_threshold_sensitivity} examines the sensitivity of this pattern to the visibility thresholds used during construction.

\section{Conclusion}

We present \benchmarkName, a construction-grounded benchmark for layer-wise chart understanding and editing. Executable chart programs provide the rendered charts and the structural information used to construct the evaluation. The understanding track directly queries layer attribution, layer binding, and visibility ordering. The editing track evaluates end-to-end transformations and reports results across local target, binding-consistent, and visibility-constrained families derived from the chart construction.

Visibility ordering is the lowest-scoring understanding dimension for all seven VLMs in the main comparison. Qwen3.5-27B, which achieves the highest macro-average accuracy, obtains 93.04\% on layer attribution and 97.46\% on layer binding, but 61.46\% on visibility ordering. Overall mIoU ranges from 1.49\% to 4.93\%. Visibility-constrained edits score lowest for every editor, with mIoU of 0.37\%--2.00\%. Across both tracks, the clearest recurring performance gap appears in tasks involving front-to-back relations between overlapping chart components.

Our study is limited to program-generated charts from the 14 paradigms supported by our construction library. Family-level editing scores reflect both operation mix and structural conditions, which may co-occur across items. The editing metrics measure agreement with reference outputs; they do not directly reveal whether a model internally represents chart layers.

Future work can extend the benchmark to naturally occurring charts and a broader range of visualization paradigms and transformations. Matched editing sets could isolate individual structural conditions and support cleaner comparisons across families. The observed weakness on tasks about visibility ordering also motivates probes of how models represent front-to-back relations and experiments with layer-aware training objectives or explicit intermediate representations.

\paragraph{Use of generative AI.}
OpenAI Codex assisted with language polishing and code implementation based on the authors' ideas. The authors take responsibility for the content of the manuscript and the accompanying implementation.

\bibliographystyle{iclr2027_conference}
\bibliography{references}

\appendix \section{Terminology}
\label{app:terminology}

Layer Attribution, Layer Binding, and Visibility Ordering are the three core layer-wise concepts formalized as evaluation dimensions in Section~\ref{sec:formulation}. We use the same labels for the corresponding question families in the understanding track. The editing track uses the separate family labels listed below.

\begin{table}[H]
\centering

\caption{Canonical terminology used throughout \benchmarkName.}
\label{tab:terminology}
\begingroup
\small
\setlength{\tabcolsep}{6pt}
\renewcommand{\arraystretch}{1.15}
\begin{tabularx}{\textwidth}{
@{}>{\raggedright\arraybackslash}p{0.22\textwidth}
>{\raggedright\arraybackslash}X@{}}
\toprule
\textbf{Term} & \textbf{Use in this paper} \\
\midrule

Layer-wise representation
& A component-level description of a chart for the evaluated tasks within our controlled construction domain, written as $S=(C,\Lambda)$. Here, $C$ contains chart components, and $\Lambda$ records their organization and relations. \\
\graydashline

Component
& A visible chart element or chart-level layout structure together with its applicable resolved data, content, appearance, and spatial properties. \\
\graydashline

Functional layer
& The operational role assigned to a component by $\lambda$: Reference, Data, Text, or Dependent. The complete ontology is given in Appendix~\ref{app:operational_layer_ontology}. \\
\graydashline

Layer Attribution
& The component-level dimension concerning the functional-layer assignment $\lambda(c)$ of a component $c$. \\
\graydashline

Layer Binding
& The relation-level dimension concerning a construction-derived semantic correspondence in $\mathcal{B}$ between components in different functional layers. \\
\graydashline

Visibility Ordering
& The relation-level dimension concerning the resolved front-to-back relation in $\mathcal{V}$ between overlapping components. \\
\graydashline

Layer-wise understanding
& The question-answering track that directly queries the three layer-wise dimensions from a rendered chart. Its three question families use the corresponding dimension labels. \\
\graydashline

Layer-wise editing
& The end-to-end track that asks a model to transform a rendered chart according to an instruction and evaluates the output under construction-derived structural conditions. \\
\graydashline

Local target edit
& An editing family for a specified target or layout edit that does not require the benchmark's binding- or visibility-specific condition. \\
\graydashline

Binding-consistent edit
& An editing family whose reference edit visibly changes both a Data target and a construction-linked Dependent component. \\
\graydashline

Visibility-constrained edit
& An editing family whose reference edit changes a target within a verified overlap while preserving the construction-resolved front-to-back order. \\

\bottomrule
\end{tabularx}
\endgroup
\end{table}

\FloatBarrier

\section{Operational Layer Ontology}
\label{app:layer_ontology}

\subsection{Functional Layer Ontology and Element Mapping}
\label{app:operational_layer_ontology}

Table~\ref{tab:terminology} summarizes the terminology used throughout the paper. \benchmarkName organizes chart elements into four functional layers: \emph{Reference}, \emph{Data}, \emph{Text}, and \emph{Dependent}. This ontology is the operational decomposition used across the 14 programmatic chart paradigms in the benchmark. Its scope is the element vocabulary instantiated by these paradigms. Elements are classified by their function in a chart, rather than by the graphical primitive or software object used to render them. Within this vocabulary, every canonical element type is assigned to exactly one layer.

\paragraph{Reference layer.}
Reference elements provide the coordinate and positional apparatus through which readers locate marks within a scale, category, named axis dimension, or plotting frame. They calibrate, index, or orient positions rather than encode primary data records. Axis titles belong to this layer because they name coordinate dimensions.

\paragraph{Data layer.}
Data elements are the marks and geometric objects that directly encode values, observations, aggregates, or distributions. The layer includes both primary marks and derived statistical marks when their geometry represents a relation in the data, as in a fitted regression line.

\paragraph{Text layer.}
The Text layer is reserved for standalone explanatory text that names, describes, or contextualizes the chart. Such text is neither part of a positional scale or index nor attached to a particular data mark or encoding channel.

\paragraph{Dependent layer.}
Dependent elements derive their interpretation from a data mark, mark group, or encoding channel. They decode an encoding or attach information to encoded data. Legends, colorbars, size legends, data labels, and mark-tied callouts therefore remain in the Dependent layer even when they are rendered as text.

\paragraph{Functional boundary rule.}
Because the ontology is functional, visually similar objects may belong to different layers. A plotted line is Data when it represents a series or fitted relation, whereas a positional baseline, grid line, or scale-reference guide is Reference. Tick labels, categorical axis labels, heatmap row and column labels, and axis titles are Reference. Chart titles, captions, and free annotations that are not tied to a mark or encoding channel are Text. Legend entries, colorbars, data labels, and mark-tied callouts are Dependent.

\paragraph{Instantiated element assignments.}
The element classes below are standard chart components. Table~\ref{tab:element_layer_mapping} records how the functional definitions above are instantiated across the 49 canonical element classes used by the benchmark. This explicit assignment ensures that the same element class is treated consistently across source programs and benchmark instances. 
\begin{table}[H]
\centering

\caption{Operational element-to-layer mapping used by \benchmarkName. Elements are assigned according to their function in the chart.}
\label{tab:element_layer_mapping}
\begingroup
\small
\setlength{\tabcolsep}{7pt}
\renewcommand{\arraystretch}{1.15}
\begin{tabularx}{\textwidth}{
@{}>{\raggedright\arraybackslash}p{0.14\textwidth}
>{\raggedright\arraybackslash}X@{}}
\toprule
\textbf{Layer} & \textbf{Canonical element classes} \\
\midrule

Reference & axes; axis lines; axis titles (\texttt{xlabel}, \texttt{ylabel}, and \texttt{zlabel}); ticks; tick labels; categorical axis labels; heatmap row labels; heatmap column labels; grid lines; baselines; zero lines; reference lines; reference bands; plotting frames; polar or radar angular and radial guides. \\
\graydashline

Data & bars; lines; points; areas; heatmap cells; pie or rose wedges; radar or polar polygons; bubble marks; fitted regression lines; boxplot bodies; boxplot whiskers; boxplot caps; boxplot medians; outliers; histogram bins. \\
\graydashline

Text & chart titles; figure-level titles; subplot or panel titles; subtitles; panel labels; standalone annotations; captions; source notes. \\
\graydashline

Dependent & legends; legend entries; legend titles; legend symbols or handles; colorbars; colorbar tick labels; colorbar labels; size legends; data labels; mark-tied callouts; labels printed inside or next to marks. \\
\bottomrule
\end{tabularx}
\endgroup
\end{table}

\section{Benchmark Construction Details}
\label{app:construction_details}

We construct two complementary benchmark tracks from charts rendered from source programs with the retained layer-wise representation. The understanding track queries functional layer roles and structural relations in a rendered chart, whereas the editing track asks models to apply controlled modifications specified by natural-language instructions. The following subsections detail the construction of the two types of benchmark instances.

\subsection{Understanding Data Construction}
\label{app:understanding_data_construction}

Each understanding item is derived from a canonical fact constructed from the source chart program and its aligned layer renderings. The canonical fact records the fact type, visual anchor, supporting layers, and answer target independently of question wording.

\paragraph{Family Assignment.}

The fact type deterministically assigns each canonical fact to exactly one of Layer Attribution, Layer Binding, or Visibility Ordering. A source chart may produce multiple facts and therefore contribute items to different families.

\medskip
\noindent\textbf{Layer Attribution.} Layer Attribution asks which functional layer contains a specified chart element. The routing follows the operational ontology: \texttt{reference\_layer\_role} and \texttt{axis\_title\_role} map to Reference; \texttt{data\_mark\_role} maps to Data; \texttt{standalone\_text\_role} maps to Text; and \texttt{legend\_block\_role}, \texttt{colorbar\_role}, and \texttt{data\_label\_role} map to Dependent. Each item uses the four layer roles as its answer choices.

\medskip
\noindent\textbf{Layer Binding.} Layer Binding asks which semantic partner corresponds to a specified Data-layer element. We instantiate four typed relations: mark-to-legend-entry and mark-to-data-label connect Data to Dependent, while mark-to-axis-category and heatmap-cell-to-row/column-label connect Data to Reference. The source program determines the gold partner and supplies relation-compatible distractors.

\medskip
\noindent\textbf{Visibility Ordering.} Following the standard alpha-compositing model, we define Visibility Ordering by the stacking order used to composite two overlapping rendered elements \citep{porter1984compositing}. We instantiate this definition using the renderer-resolved drawing order of Matplotlib~\citep{hunter2007matplotlib}.

Visibility Ordering asks which of two locally overlapping elements is rendered in front. All layer renderings used here are exported from the same Matplotlib figure at 100\,dpi and therefore share a common pixel coordinate system. For layer assets \(i\) and \(j\), let \(A_i=\{p\mid\alpha_i(p)>8\}\) and \(n_{ij}=|A_i\cap A_j|\). A pair is retained when
\begin{equation}
    n_{ij}\geq80,
    \qquad
    \frac{n_{ij}}{\max\!\left(1,\min\!\left(|A_i|,|A_j|\right)\right)}
    \geq0.015.
\end{equation}
The benchmark items cover overlapping area fills, radar fills, bubbles, scatter points, and Bar3D series. For two-dimensional charts, the later element in the recorded artist render order is treated as frontmost. For Bar3D, frontmost order is resolved after drawing the chart under the fixed camera configuration \(\mathrm{elev}=24^\circ\), \(\mathrm{azim}=-58^\circ\), and \(\mathrm{roll}=0^\circ\).

\paragraph{Question verbalization.}
After the canonical fact and answer space are fixed, we use \texttt{qwen3.6-plus} to verbalize each item as a concise, image-grounded multiple-choice question. Figure~\ref{fig:understanding-verbalization-prompt} shows the shared system and user prompt templates.

\begin{figure}[!t]
\centering
\begin{tcolorbox}[
  enhanced,
  width=\linewidth,
  colback=white,
  colframe=black,
  colbacktitle=black,
  coltitle=white,
  title={Prompt for Question Verbalization},
  fonttitle=\ttfamily\bfseries,
  fontupper=\ttfamily\footnotesize,
  boxrule=0.65pt,
  arc=0pt,
  left=1.7mm,
  right=1.7mm,
  top=1.2mm,
  bottom=1.2mm,
  before upper={
    \raggedright
    \setlength{\parindent}{0pt}
  }
]
\textbf{SYSTEM}\par
\smallskip

You are writing one chart QA item from a fixed canonical fact.\par
Do not change the answer, family, or distractor set.\par
Do not invent evidence or hidden objects.\par
Do not use forbidden strings.\par
Write a short, chart-specific, image-grounded question that matches the
assigned question angle and style.\par
Return valid JSON only.\par

\medskip
\textbf{USER}\par
\smallskip

Write exactly one multiple-choice chart QA question.\par

\smallskip
\textbf{Requirements:}\par
- The answer is already fixed by the canonical fact.\par
- You must preserve the assigned family and question angle.\par
- The question must require looking at the chart image.\par
- The question must not reveal the answer text or forbidden strings.\par
- The question must be short and concrete.\par
- The final choices are already locked; do not rewrite or reorder them.
\end{tcolorbox}

\caption{
Shared system and user prompt fragments used for question verbalization.
Instance-specific structured context is omitted for clarity.
}
\label{fig:understanding-verbalization-prompt}
\end{figure}

\subsection{Editing Variant Construction}
\label{app:editing_data_construction}

Each editing item pairs an instruction with aligned input and ground truth renderings produced from the same chart program. The structured parameter modification and aligned layer exports are construction metadata. Editing models receive only the input chart and instruction.

\begin{figure}[!t]
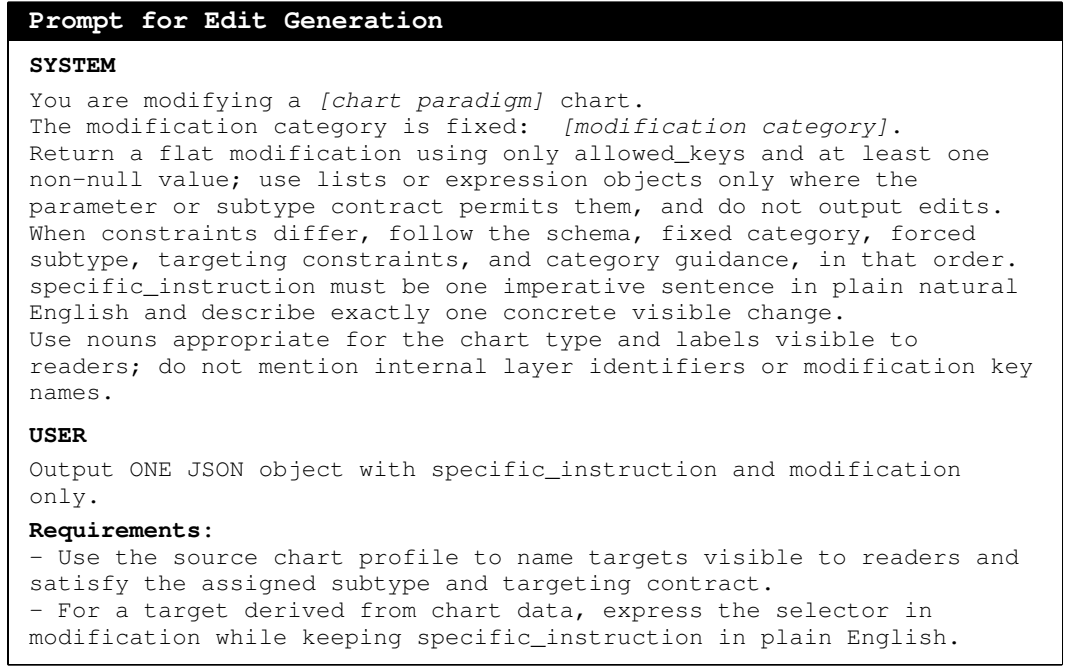

\centering
\begin{tcolorbox}[
  enhanced,
  width=\linewidth,
  colback=white,
  colframe=black,
  colbacktitle=black,
  coltitle=white,
  title={Prompt for Edit Generation},
  fonttitle=\ttfamily\bfseries,
  fontupper=\ttfamily\footnotesize,
  boxrule=0.65pt,
  arc=0pt,
  left=1.7mm,
  right=1.7mm,
  top=1.2mm,
  bottom=1.2mm,
  before upper={
    \raggedright
    \setlength{\parindent}{0pt}
  }
]
\textbf{SYSTEM}\par
\smallskip
You are modifying a \emph{[chart paradigm]} chart.\par
The modification category is fixed: \emph{[modification category]}.\par
Return a flat \texttt{modification} using only \texttt{allowed\_keys} and at least one non-null value; use lists or expression objects only where the parameter or subtype contract permits them, and do not output \texttt{edits}.\par
When constraints differ, follow the schema, fixed category, forced subtype, targeting constraints, and category guidance, in that order.\par
\texttt{specific\_instruction} must be one imperative sentence in plain natural English and describe exactly one concrete visible change.\par
Use nouns appropriate for the chart type and labels visible to readers; do not mention internal layer identifiers or modification key names.

\medskip
\textbf{USER}\par
\smallskip
Output ONE JSON object with \texttt{specific\_instruction} and \texttt{modification} only.

\smallskip
\textbf{Requirements:}\par
- Use the source chart profile to name targets visible to readers and satisfy the assigned subtype and targeting contract.\par
- For a target derived from chart data, express the selector in \texttt{modification} while keeping \texttt{specific\_instruction} in plain English.
\end{tcolorbox}
\caption{Abridged and reformatted reconstruction of the shared prompt contract for canonical seed generation with Qwen3.6-Plus. Structured blocks for individual instances are omitted, and no chart image is supplied.}
\label{fig:editing_generation_prompt}
\end{figure}

\paragraph{Instruction and modification generation.}
For each canonical seed, the pipeline fixes the chart paradigm, a modification category (\emph{Data-centric}, \emph{Visual}, \emph{Text/Font}, or \emph{Layout/Margin}), a subtype, and its targeting contract. The category constrains parameter generation but does not determine the editing family. The parameter schema intersects the editable arguments of the rendering function with the selected category partition, excluding \texttt{edits}, identity controls, and global structure controls. We use \texttt{qwen3.6-plus} with the text prompt in Fig.~\ref{fig:editing_generation_prompt}; no chart image is provided. The prompt supplies the source chart profile, schema, category guidance, subtype contract, and targeting constraints. It requests an imperative \texttt{specific\_instruction} and a flat parameter map with at least one non-null allowed key. Lists or expression objects are permitted only by the relevant parameter or subtype contract. Expression subtypes require an expression on the designated target key that references an allowed chart context. After key validation and expression evaluation, eligible numeric data values are projected or clamped when necessary. The resolved modification is merged with the original arguments and rendered by the same chart function and layer exporter.

\paragraph{Restricted parameter updates.}
For each canonical edit, the runner derives an allowed parameter set from the corresponding builder signature. It removes \texttt{edits}, fields that determine sample identity or global structure, and a small number of paradigm-specific fields, then restricts the remainder to the assigned Data-centric, Visual, Text/Font, or Layout/Margin category. Recognized schema wrappers and aliases are normalized. The \texttt{modification} must contain at least one non-null field; every such field must belong to the active category schema, and the update must satisfy the required fields, permitted alternatives, target dependencies, and forbidden fields of its assigned subtype. Symbolic values are Python expressions evaluated with \texttt{eval} in a reduced namespace containing NumPy, chart data, and a small helper set. Eligible data mutations are checked against paradigm-specific rendering bounds and, when possible, deterministically projected back into range; any unresolved case is recorded. After excluded fields are removed, the resolved update is overlaid on the original parameters, conflicting stale mutation, targeting, or style fields are cleared, and the final arguments are filtered by the builder signature.

\paragraph{Deterministic re-execution.}
Once a proposed edit has been resolved to a concrete parameter update, each reference edit is reconstructed independently by applying that update to the stored source parameters and original random seed, rather than to a previously edited image or parameter state. No additional model call is made during reconstruction. The resolved update and stored source parameters fully specify the builder inputs. Our reproducibility claim assumes fixed dependency versions and fonts; it does not extend to byte-identical rendering across Matplotlib versions or platform font stacks.

\paragraph{Variant expansion.}
Each accepted seed is expanded deterministically into up to five scaled variants by changing feasible values while preserving the target, subtype, and modification keys. A variant is materialized only when rendering succeeds without repair; subsequent instruction and profile audits determine whether it is eligible for evaluation.

\paragraph{Family Assignment.}

Family membership describes the relational demand among rendered elements in the reference edit, rather than its surface modification category. An offline assignment stage routes each item to a Local target edit, Binding-consistent edit, or Visibility-constrained edit using its subtype, dependencies, changed layer units, and the relation rules of the chart paradigm. When both relation types apply, the visibility route takes precedence. Verification after rendering checks only the selected route. Insufficient visible activity in valid regions assigns the item to Local target edit, while missing renderings, layer assets, or required masks leave the item without a family label. Assignment and verification use only construction metadata, renderer exports, and the input and reference renderings; model predictions are used only during evaluation.

\medskip
\noindent\textbf{Local target edit.}
Local target edits include edits routed directly to this family and relational edits that fail verification despite having valid regions. It is not tied to any one modification category.

\medskip
\noindent\textbf{Binding-consistent edit.}
Binding-consistent edits are edits for which construction metadata selects a Layer Binding relation between a Data target and a Dependent component. Let $I$ and $G$ be the aligned 8-bit input and ground truth RGB renderings. We define the visible change mask and each exported layer's support mask as
\begin{equation}
\begin{aligned}
    V(p)&=\mathbf{1}\!\left[
    \frac{1}{3}\sum_{c=1}^{3}\lvert G_c(p)-I_c(p)\rvert \geq 3
    \right],\\
    M_{\ell}(p)&=\mathbf{1}[\alpha_{\ell}(p)>8],
\end{aligned}
\end{equation}
where $\alpha_{\ell}$ is the 8-bit alpha channel of the corresponding layer export for the reference edit. All set cardinalities below count pixels, and the thresholds are frozen construction constants. These family verification masks are separate from the headline editing metrics. A proxy mask $T$ unions Data groups selected by changed unit or path indices; if this mask is empty, it falls back to all Data exports. A second mask $D$ is the union of eligible Dependent exports for legends, colorbars, size legends, and data labels. The route is retained as a Binding-consistent edit only if
\begin{equation}
    \lvert T\cap V\rvert\geq16
    \quad\text{and}\quad
    \lvert D\cap V\rvert\geq16.
\end{equation}
Metadata supplies the semantic relation. Supported routes link mark appearance or encoding to legends or colorbars, and mark size to size legends. The masks only confirm visible change on both sides.

\medskip
\noindent\textbf{Visibility-constrained edit.}
Visibility-constrained edits are edits for which construction metadata selects a Visibility Ordering relation involving the edited target. It reuses $V$, $M_{\ell}$, and the Data target proxy $T$. Data exports are grouped using normalized filenames, while eligible Text and Dependent exports enter as individual candidates. For each pair $(i,j)$, the verifier defines
\begin{equation}
\begin{aligned}
    R_{ij}&=M_i\cap M_j,\\
    O_{ij}&=
    \begin{cases}
        R_{ij}\cap T, & \lvert T\rvert>0,\\
        R_{ij}, & \lvert T\rvert=0.
    \end{cases}
\end{aligned}
\end{equation}
Among pairs with $\lvert O_{ij}\rvert\geq32$, the largest candidate region is selected as $O$, with ties following sorted pair order. The route is retained as a Visibility-constrained edit only if
\begin{equation}
    \lvert O\cap V\rvert\geq16.
\end{equation}
Metadata supplies the visibility relation. This test confirms visible change in an eligible overlap; it does not identify the foreground element, verify direction, or establish that stacking order changed.

\subsection{Construction Validation and Dataset Accounting}
\label{app:construction_validation}

The two benchmark tracks use validation procedures matched to how their supervision is constructed. All checks operate on source programs, construction records, and reference renderings; outputs from the evaluated models are not used to select benchmark items.

\paragraph{Understanding.}
Before verbalization, each item is checked for a valid canonical fact, an eligible task family, existing visual evidence, and a unique gold answer among the locked choices. After verbalization, deterministic checks verify the output schema, preserve the assigned family and answer, and reject answer leakage or duplicated choices. We additionally conduct a family-stratified audit of 100 items assisted by \texttt{qwen3.6-plus} sampled from the full 7,329-item set, assessing gold-answer correctness, visual answerability, and answer uniqueness. Ninety-nine items passed this audit before a wording correction to the remaining item. The 20 Bar3D Visibility Ordering questions in the test set are separately checked after resolving their front-to-back relations under the fixed camera configuration, and all pass.

\paragraph{Editing.}
Each candidate variant must pass two construction checks. Profile verification reconstructs the source and edited chart profiles and tests whether the resolved parameter update produces the intended change. Instruction alignment checks whether the natural-language instruction describes the same target and modification encoded by that update. Numerical alignment accepts rounded forms of the stored values. Some instructions round numerical values while reference edits retain the stored precision, so passing this check does not guarantee exact numerical agreement between the wording and the reference. Of the 5,392 held-out candidates, 5,148 pass both checks. We exclude 514 no-op reference edits and 20 additional candidates lacking the layer assets required for family assignment, producing the common 4,614-item scored set.

\paragraph{Instruction precision.}
A check of instructions with the template ``Multiply every point on \ldots\ by \ldots'' identified 45 scored items whose displayed multipliers differ from the stored values. Recomputing mIoU from the saved per-item scores after excluding these items changes overall mIoU by at most 0.017 percentage points across the four editors. Their overall ranking is unchanged, and the Visibility-constrained family remains lowest for each editor. This check covers that instruction template and does not exhaustively assess all numerical instructions.

\section{Inference Settings}
\label{app:inference_settings}

\paragraph{Understanding.}
All nine vision-language models are evaluated without sampling, with a maximum of 32 generated tokens and a fixed seed of 0. Models run in bfloat16 without quantization, using model-specific chat formatting and image preprocessing. For Qwen3.5, the wrapper passes \texttt{enable\_thinking=False} when supported by the chat-template interface. The Qwen3-VL model uses the \path{Qwen/Qwen3-VL-8B-Instruct} checkpoint.

The shared task prompt contains the four functional layer definitions, the question, and the locked answer choices. The image+text condition also supplies the rendered chart. In the text-only condition, the image is omitted and the sentence \texttt{No chart image is provided in this control condition.} is inserted after \texttt{Answer the chart question by choosing one option.}

\paragraph{Editing.}
Table~\ref{tab:editing_inference_settings} lists the settings used for the four image editors. We generate one output per item with a fixed seed of 20260530. InstructPix2Pix uses the Euler ancestral scheduler supplied with the checkpoint. OmniGen2 applies classifier-free guidance over the full denoising interval, and Qwen Image Edit 2509 uses a blank negative prompt. InstructPix2Pix and OmniGen2 generate at the input resolution; FLUX Kontext Dev and Qwen Image Edit 2509 use their native output sizes, followed by the evaluation normalization described in Appendix~\ref{app:editing_evaluation}.

\begin{table}[H]
    \centering
    \caption{Inference settings for the evaluated image editors. Text/CFG denotes guidance scale for FLUX Kontext Dev and InstructPix2Pix, text guidance for OmniGen2, and true CFG for Qwen Image Edit 2509.}
    \label{tab:editing_inference_settings}
    \begingroup
    \small
    \setlength{\tabcolsep}{4pt}
    \renewcommand{\arraystretch}{1.08}
    \begin{tabularx}{\linewidth}{@{}lXrrrr@{}}
        \toprule
        \textbf{Editor} & \textbf{Checkpoint} & \textbf{Steps} & \textbf{Text/CFG} & \textbf{Image CFG} & \textbf{Precision} \\
        \midrule
        FLUX Kontext Dev & \path{black-forest-labs/FLUX.1-Kontext-dev} & 28 & 2.5 & -- & bfloat16 \\
        InstructPix2Pix & \path{timbrooks/instruct-pix2pix} & 100 & 7.5 & 1.5 & float16 \\
        OmniGen2 & \path{OmniGen2/OmniGen2} & 50 & 5.0 & 2.0 & bfloat16 \\
        Qwen Image Edit 2509 & \path{Qwen/Qwen-Image-Edit-2509} & 40 & 4.0 & -- & bfloat16 \\
        \bottomrule
    \end{tabularx}
    \endgroup
\end{table}

\section{Evaluation Metrics}
\label{app:evaluation_metrics}

\subsection{Understanding Evaluation}
\paragraph{Answer parsing and scoring.}
We evaluate understanding on the 727-item test split, which contains 316 Layer Attribution, 315 Layer Binding, and 96 Visibility Ordering items. After whitespace normalization, a deterministic parser attempts to map each model response to one of the locked choices. It examines structured prediction fields before the raw response. For each candidate string, locked choice identifiers and texts are considered in decreasing surface length. The parser first attempts an exact match or a match at the beginning of the response, followed by substring matching. The inference wrapper also extracts the first standalone A--D token from a longer response and interprets it according to the displayed choice order. Bare numerals are not interpreted as choice positions.

Every accepted run must contain exactly one result record for each test item. Blank, duplicate, missing, or extra item identifiers invalidate the run. Within a complete run, request failures, empty responses, and responses that cannot be mapped to a locked choice remain in the denominator and are scored as incorrect.

Let \(\mathcal{I}_f\) contain the items in family \(f\), let \(N_f=|\mathcal{I}_f|\), and let \(y_i\) and \(\widehat{y}_i\) denote the gold and parsed choices for item \(i\). We set \(\widehat{y}_i=\bot\) when the response cannot be mapped to a locked choice. With \(s_i\) denoting the request status, family accuracy is
\begin{equation}
    \operatorname{Acc}_f
    =
    \frac{1}{N_f}
    \sum_{i\in\mathcal{I}_f}
    \mathbf{1}\!\left[
        s_i=\mathrm{ok}
        \land
        \widehat{y}_i=y_i
    \right].
\end{equation}

The reported Macro Average gives equal weight to Layer Attribution, Layer Binding, and Visibility Ordering:
\begin{equation}
    \operatorname{MacroAcc}
    =
    \frac{1}{3}
    \sum_{f\in\{\mathrm{attr},\mathrm{bind},\mathrm{vis}\}}
    \operatorname{Acc}_f.
\end{equation}

Accuracy over all 727 items is computed using the same correctness rule over the complete test set and therefore weights every item equally. We also report the invalid choice rate:
\begin{equation}
    \operatorname{InvalidRate}
    =
    \frac{1}{727}
    \sum_{i=1}^{727}
    \mathbf{1}\!\left[
        \widehat{y}_i=\bot
    \right].
\end{equation}
All accuracies and rates are reported as percentages.

\subsection{Editing Evaluation}
\label{app:editing_evaluation}

For item $i$, let $I_i$, $G_i$, and $P_i$ denote the input, reference edit, and normalized prediction. All PaintBench and SSIM results reported in this paper use the common scored set of 4,614 items: 3,447 Local target edits, 444 Binding-consistent edits, and 723 Visibility-constrained edits. OCR is evaluated separately on 568 eligible items among 581 applicable \texttt{Text/Font} edits. A missing or invalid prediction image invalidates the evaluation rather than reducing its denominator.

\paragraph{Image normalization.}
The input and raw prediction are corrected for EXIF orientation and converted to RGB. If their resolutions match, the prediction is used without resampling. Otherwise, for input resolution $(W,H)$ and prediction resolution $(w,h)$, we compute
\begin{equation}
    \epsilon=\frac{|w/h-W/H|}{W/H}.
\end{equation}
When $\epsilon\leq0.03$, Lanczos resizing covers $(W,H)$ followed by a centered crop. Otherwise, Lanczos resizing fits the prediction inside a $W\times H$ canvas, followed by centered padding. Resized dimensions are rounded to the nearest integer with ties to even, and center offsets use integer floor division. Each padding channel is the third value after sorting that channel over the four input corners. The resulting $P_i$ matches the dimensions of $I_i$ and $G_i$, with no further resizing or alignment during scoring.

\paragraph{PaintBench metrics.}
Following PaintBench~\citep{xu2026paintbench}, we convert the aligned 8-bit RGB prediction and reference from IEC sRGB to D65 CIE Lab using \texttt{float32} arithmetic and compute their per-pixel CIE76 distance $d_i(x)$. Let $\Omega_i$ be the image domain and define
\begin{equation}
    E_i=\{x\in\Omega_i:\exists c,\ I_i(x,c)\neq G_i(x,c)\},
    \qquad
    U_i=\Omega_i\setminus E_i,
    \qquad
    C_{i,t}=\{x\in\Omega_i:d_i(x)\leq t\},
\end{equation}
where $t\in\{0,\ldots,10\}$. We compute
\begin{equation}
\begin{aligned}
    \operatorname{EditAcc}_{i,t}
        &=\frac{|E_i\cap C_{i,t}|}{|E_i|},\\
    \operatorname{PresAcc}_{i,t}
        &=\frac{|U_i\cap C_{i,t}|}{|U_i|},\\
    \operatorname{IoU}^{\mathrm{PB}}_{i,t}
        &=\frac{|E_i\cap C_{i,t}|}
        {|E_i|+|U_i\setminus C_{i,t}|}.
\end{aligned}
\end{equation}
The scored set has $|E_i|>0$; Preservation Accuracy is defined as one when $|U_i|=0$. For any of the three metrics $m$ and reported item set $\mathcal{D}$, either the overall set or one family,
\begin{equation}
    \operatorname{Score}_{\mathcal{D}}(m)
    =
    \frac{1}{|\mathcal{D}|}
    \sum_{i\in\mathcal{D}}
    \frac{1}{11}
    \sum_{t=0}^{10}m_{i,t}.
\end{equation}
PaintBench mIoU is the IoU defined above, not IoU between binary change masks. All three metrics are reported as percentages, with higher values being better.

\paragraph{Structural similarity.}
We compute RGB SSIM following \citet{wang2004ssim}, using Gaussian weights with $\sigma=1.5$, an $11\times11$ window, population covariance, $K_1=0.01$, $K_2=0.03$, and data range 255. Scores are averaged equally over valid spatial positions and RGB channels. For each item,
\begin{equation}
    S_{0,i}=\operatorname{SSIM}(I_i,G_i),
    \qquad
    S_i=\operatorname{SSIM}(P_i,G_i),
    \qquad
    \Delta S_i=S_i-S_{0,i}.
\end{equation}
The three reported values are unweighted means over the same 4,614 items and are multiplied by 100 without clipping or further normalization. Higher $S_i$ is better, while $\Delta S_i>0$ indicates improvement over the input baseline. Because SSIM measures global similarity, it is interpreted together with this baseline and the PaintBench metrics.

\paragraph{OCR text preservation.}
For each applicable item, we form
\begin{equation}
    M_i=\{x\in\Omega_i:\max_c|I_i(x,c)-G_i(x,c)|>8\},
\end{equation}
take its bounding box, expand it by eight pixels on each side, and clip it to the image boundary. OCR is applied to the input and reference crops at clockwise rotations $0^\circ$, $90^\circ$, $180^\circ$, and $270^\circ$. Segments are joined in engine order and normalized using Unicode NFKC and whitespace collapse, while case, digits, and punctuation are preserved.

A rotation is eligible when the input and reference produce the same nonempty string. If several rotations qualify, we maximize, in order, the smaller of the two mean confidences, their average, and the normalized text length, with the earlier rotation in the fixed order preferred on any remaining tie. The selected crop, rotation, and reference string are shared by all models.

Let $r_i$ and $p_i$ be the normalized reference and prediction strings, and let $d$ be their character Levenshtein distance. We report
\begin{equation}
    \operatorname{Exact}_i
    =
    \mathbf{1}[r_i=p_i],
    \qquad
    \operatorname{OCR\text{-}NES}_i
    =
    1-\frac{d(r_i,p_i)}
    {\max(|r_i|,|p_i|)}.
\end{equation}
The shared rule retains 568 of 581 items for every model. An empty OCR output string remains in the denominator and receives zero for both metrics; a backend failure invalidates the evaluation. Exact and OCR-NES are unweighted means over these 568 items and are reported as percentages. OCR-NES measures preservation of readable text, not correctness of the requested font or style change.

We use \texttt{PP-OCRv6\_medium\_det} and \texttt{PP-OCRv6\_medium\_rec}~\citep{zhang2026ppocrv6} with PaddleOCR 3.7.0, PaddlePaddle 3.3.1, and PaddleX 3.7.2 on CPU. MKL-DNN, document orientation classification, document unwarping, and text line orientation are disabled.

\section{Additional Experimental Results}
\label{app:additional_results}

\subsection{Complementary Understanding Results}
\label{app:additional_understanding_results}

\paragraph{Text-only control.}
To assess the role of the chart image, we evaluate the same seven VLMs on all 727 test items without images, retaining the four functional layer definitions, questions, and answer choices. The text-only prompt additionally states that no chart image is provided. Generation settings, answer parsing, and scoring are unchanged, with invalid outputs counted as incorrect. Table~\ref{tab:qa_text_only_control} reports the results.

\begin{table}[H]
    \centering
    \small
    \begingroup
    \setlength{\tabcolsep}{3pt}
    \renewcommand{\arraystretch}{1.08}
    \begin{tabular}{@{}lrrrrr@{}}
        \toprule
        \textbf{Model}
        & \shortstack{\textbf{Macro}\\\textbf{Average} $\uparrow$}
        & \shortstack{\textbf{Layer}\\\textbf{Attribution} $\uparrow$}
        & \shortstack{\textbf{Layer}\\\textbf{Binding} $\uparrow$}
        & \shortstack{\textbf{Visibility}\\\textbf{Ordering} $\uparrow$}
        & \textbf{Invalid} $\downarrow$ \\
        \midrule
        Qwen3.5-2B
        & 58.81 & 78.48 & 55.24 & 42.71 & 0.14 \\
        Qwen3.5-9B
        & 59.13 & 87.97 & 42.54 & 46.88 & 2.06 \\
        Qwen3.5-27B
        & \textbf{64.13} & 84.18 & \textbf{71.75} & 36.46 & \textbf{0.00} \\
        Qwen3-VL-8B
        & 57.68 & 87.66 & 37.46 & 47.92 & \textbf{0.00} \\
        InternVL3-8B
        & 60.36 & 87.03 & 26.35 & \textbf{67.71} & 10.45 \\
        InternVL3-14B
        & 61.19 & 90.19 & 37.14 & 56.25 & 4.13 \\
        MiniCPM-V-4.5
        & 54.62 & \textbf{92.41} & 26.67 & 44.79 & 15.68 \\
        \midrule
        Uniform random
        & 34.80 & 25.00 & 29.39 & 50.00 & -- \\
        \bottomrule
    \end{tabular}
    \endgroup
    \caption{
    Text-only control on the 727-item understanding test set.
    The chart image is removed while the question and answer choices are retained.
    All values are percentages.
    Macro Average is the unweighted average of the three family accuracies.
    Bold marks the best model result in each column.
    }
    \label{tab:qa_text_only_control}
\end{table}

All seven VLMs have lower macro-average accuracy in the text-only condition, with decreases of 10.16--25.77 percentage points relative to Table~\ref{tab:qa_main}. Layer Attribution nevertheless remains at 78.48--92.41\%, showing that many questions about functional roles can be answered without the image. For Layer Binding, Qwen3.5-27B scores 71.75\% without images and 97.46\% with images, with no invalid outputs in either condition. Some of the larger Binding gaps also involve output failures. MiniCPM-V-4.5 and InternVL3-8B produce 114 and 73 invalid Binding responses in the text-only control, respectively.

Visibility Ordering ranges from 36.46\% to 67.71\% without images and from 50.00\% to 64.58\% with images. In particular, InternVL3-8B scores 67.71\% in the text-only condition, compared with 56.25\% with the image. Image availability therefore does not uniformly translate into better performance on this family.

\paragraph{Runs with high invalid choice rates.}
The main image+text comparison applies an invalid choice rate ceiling of 5\%. Table~\ref{tab:qa_high_invalid} reports Qwen3.5-0.8B and Qwen3.5-4B, whose image+text runs exceed this threshold, together with their text-only controls. Every run covers all 727 test items, with invalid responses retained in the denominators and counted as incorrect.

\begin{table}[H]
    \caption{Additional understanding results for Qwen3.5-0.8B and Qwen3.5-4B on the 727-item test set. All values are percentages. Macro Average is the unweighted average of the three family accuracies.}
    \label{tab:qa_high_invalid}
    \centering
    \small
    \begingroup
    \setlength{\tabcolsep}{3pt}
    \begin{tabular}{lrrrrr}
        \toprule
        \textbf{Model}
        & \shortstack{\textbf{Macro}\\\textbf{Average} $\uparrow$}
        & \shortstack{\textbf{Layer}\\\textbf{Attribution} $\uparrow$}
        & \shortstack{\textbf{Layer}\\\textbf{Binding} $\uparrow$}
        & \shortstack{\textbf{Visibility}\\\textbf{Ordering} $\uparrow$}
        & \textbf{Invalid} $\downarrow$ \\
        \midrule
        \multicolumn{6}{l}{\textit{Image+text}} \\
        Qwen3.5-0.8B & 58.18 & 43.99 & 81.59 & 48.96 & 13.20 \\
        Qwen3.5-4B   & 65.58 & 87.03 & 81.59 & 28.13 & 7.98 \\
        \midrule
        \multicolumn{6}{l}{\textit{Text-only}} \\
        Qwen3.5-0.8B & 43.23 & 62.97 & 29.21 & 37.50 & 1.79 \\
        Qwen3.5-4B   & 56.96 & 81.01 & 37.78 & 52.08 & 0.14 \\
        \bottomrule
    \end{tabular}
    \endgroup
\end{table}

Qwen3.5-0.8B and Qwen3.5-4B produce 96 and 58 invalid responses with images, compared with 13 and 1 without images. In particular, 51 of Qwen3.5-4B's 96 Visibility Ordering responses are invalid. Its 28.13\% accuracy in this family therefore reflects both incorrect choices and frequent failures to return a valid choice.

\subsection{Complementary Editing Results}
\label{app:additional_editing_results}

Table~\ref{tab:editing_secondary_metrics} reports the complementary pixel agreement and image similarity metrics.

\begin{table}[H]
    \caption{Complementary editing results on the common set of 4,614 items. Edit Accuracy and Preservation Accuracy are reported as percentages, and SSIM is multiplied by 100. $\Delta\mathrm{SSIM}$ is the difference from the shared no-edit SSIM baseline of 95.80, reported in points and computed before rounding. Higher is better, and bold denotes the best result in each column.}
    \label{tab:editing_secondary_metrics}
    \centering
    \begingroup
    \small
    \setlength{\tabcolsep}{3pt}
    \renewcommand{\arraystretch}{1.08}
    \begin{tabular}{@{}lrrrr@{}}
        \toprule
        \textbf{Editor} & \textbf{Edit Acc.} & \textbf{Pres. Acc.} &
        \textbf{SSIM} & \textbf{$\Delta$SSIM} \\
        \midrule
        FLUX Kontext Dev & \textbf{21.91} & 78.83 & 74.84 & -20.95 \\
        InstructPix2Pix & 6.29 & 54.07 & \textbf{87.72} & \textbf{-8.07} \\
        OmniGen2 & 17.29 & 77.53 & 80.43 & -15.36 \\
        Qwen Image Edit 2509 & 18.43 & \textbf{86.23} & 78.64 & -17.15 \\
        \bottomrule
    \end{tabular}
    \endgroup
\end{table}

\paragraph{Pixel agreement and image similarity.}

FLUX Kontext Dev has the highest Edit Accuracy, while Qwen Image Edit 2509 has the highest Preservation Accuracy. InstructPix2Pix has the highest SSIM and the smallest decrease from the baseline, even though its Edit Accuracy is the lowest. This contrast shows why full-image similarity should be considered together with agreement within the edited region. All four editors remain below the no-edit SSIM baseline.

\paragraph{OCR string agreement.}

\begin{table}[H]
    \caption{OCR string agreement on the common set of 568 eligible items drawn from 581 applicable Text/Font edits. OCR Exact and OCR normalized edit similarity (OCR-NES) are reported on a 0--100 scale. Higher is better, and bold denotes the best result in each column.}
    \label{tab:editing_ocr_results}
    \centering
    \begingroup
    \small
    \setlength{\tabcolsep}{8pt}
    \renewcommand{\arraystretch}{1.08}
    \begin{tabular}{@{}lrr@{}}
        \toprule
        \textbf{Editor} & \textbf{OCR Exact} & \textbf{OCR-NES} \\
        \midrule
        FLUX Kontext Dev & 1.23 & 12.45 \\
        InstructPix2Pix & 6.69 & 49.21 \\
        OmniGen2 & \textbf{43.31} & 49.29 \\
        Qwen Image Edit 2509 & 36.09 & \textbf{57.03} \\
        \bottomrule
    \end{tabular}
    \endgroup
\end{table}

Table~\ref{tab:editing_ocr_results} reports OCR agreement on the eligible text-editing subset. OmniGen2 has the highest OCR Exact score, while Qwen Image Edit 2509 has the highest OCR-NES. On this common set, OmniGen2 more often matches the complete reference string, whereas Qwen Image Edit 2509 attains higher average string similarity when partial matches receive credit.

\section{Sensitivity to Visibility Thresholds}
\label{app:visibility_threshold_sensitivity}

We test how the reported family results change under alternative thresholds used to instantiate visibility cases. Model outputs remain fixed; only the relevant item selection or family assignment is recomputed.

\paragraph{Understanding.}
A Visibility Ordering question is retained when its overlap area is at least $p$ pixels and the overlap covers at least a fraction $r$ of the smaller component. The default values are $p=80$ and $r=0.015$, while visible support is defined using the fixed alpha cutoff $\alpha>8$. We vary one overlap threshold at a time, using $p\in\{80,90,100,110,120\}$ with $r=0.015$, and $r\in\{0.015,0.0175,0.020,0.0225,0.025\}$ with $p=80$. We report Qwen3.5-27B, the strongest evaluated model by macro-average accuracy in the main Understanding results.

\begin{figure*}[t]
    \centering
    \includegraphics[
        width=\textwidth
    ]{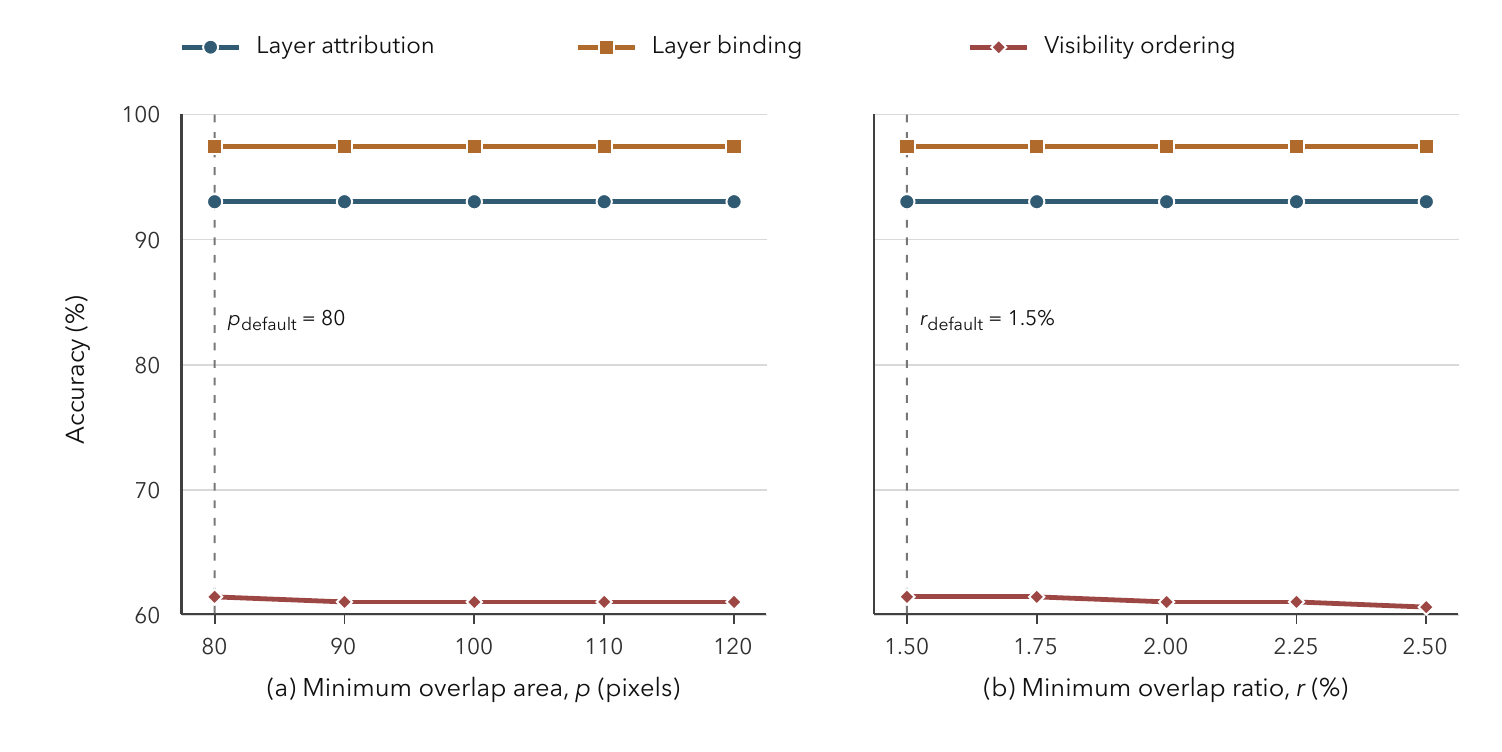}
    \caption{
        Sensitivity of Understanding accuracy to the overlap thresholds used for Visibility Ordering questions. Each panel varies one threshold while fixing the other at its default value. The dashed lines mark the defaults. The Visibility Ordering subset contains 94--96 test questions across the sweep.
    }
    \label{fig:understanding_visibility_threshold_sensitivity}
\end{figure*}

Figure~\ref{fig:understanding_visibility_threshold_sensitivity} shows that Visibility Ordering accuracy remains between 60.64\% and 61.46\%. Layer Attribution and Layer Binding stay at 93.04\% and 97.46\%, respectively, because their item sets do not depend on the overlap thresholds. The ordering of the three families is therefore unchanged across the tested stricter overlap criteria.

\paragraph{Editing.}
A candidate is assigned to the Visibility-constrained family when the selected overlap contains at least $k$ active pixels, with $k=16$ used by default. We vary $k\in\{8,12,16,20,24,28,32\}$. Candidates that fall below the cutoff are reassigned to the Local target family, while Binding-consistent assignments and the 4,614-item analysis pool remain fixed. The family mIoU values are then recomputed from the existing item scores.

\begin{figure}[t]
    \centering
    \includegraphics[
        width=\linewidth
    ]{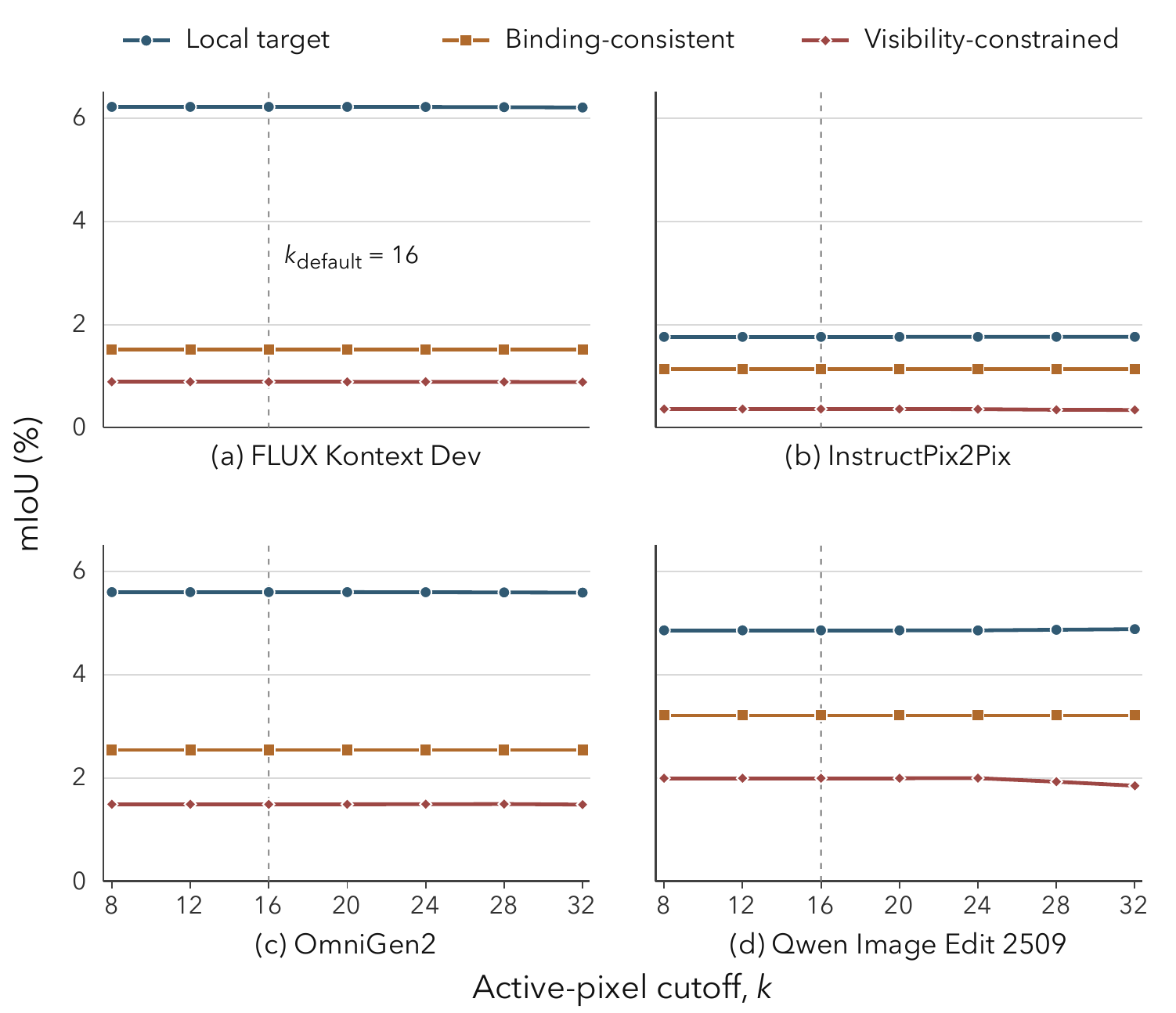}
    \caption{
        Sensitivity of editing mIoU to the active-pixel cutoff $k$. Each panel reports the three editing families for one editor after recomputing family assignments at each cutoff. The dashed lines mark the default $k=16$.
    }
    \label{fig:editing_visibility_threshold_sensitivity}
\end{figure}

As shown in Figure~\ref{fig:editing_visibility_threshold_sensitivity}, the Visibility-constrained family has the lowest mIoU for every editor at every tested cutoff. The largest change among the displayed curves is 0.15 percentage points. From $k=8$ to $k=32$, the number of Visibility-constrained edits decreases from 723 to 714, with the nine reassigned cases entering the Local target family; the Binding-consistent family remains at 444 items. Thus, the family-level pattern reported in the main evaluation is stable within the tested range of $k$.

\end{document}